\documentclass{article}

\usepackage{graphicx}
\usepackage{cite}
\usepackage{flushend}
\usepackage[utf8]{inputenc}

\usepackage{arxiv}

\usepackage[utf8]{inputenc} % allow utf-8 input
\usepackage[T1]{fontenc}    % use 8-bit T1 fonts
\usepackage{hyperref}       % hyperlinks
\usepackage{url}            % simple URL typesetting
\usepackage{booktabs, threeparttable}       % professional-quality tables
\usepackage{amsfonts}       % blackboard math symbols
\usepackage{nicefrac}       % compact symbols for 1/2, etc.
\usepackage{microtype}      % microtypography
\usepackage{lipsum}

\newcommand{\pIP}{$ I_{p} $}
\newcommand{\pVLOOP}{$ V_{loop} $}
\newcommand{\pHA}{$ H_{\alpha} $}
\newcommand{\pBOLO}{$ bolo $}
\newcommand{\pHXR}{\emph{HXR}}
\newcommand{\pSXR}{\emph{SXR}}
\newcommand{\pMIRNOV}{$ Mirnov_{16} $}
\newcommand{\pC}{$ C_{3} $}
\newcommand{\pO}{$ O_{1} $}
\newcommand{\pDELTA}{$ delta $}

\usepackage{tikz,xcolor}

\definecolor{lime}{HTML}{A6CE39}
\DeclareRobustCommand{\orcidicon}{%
	\begin{tikzpicture}
	\draw[lime, fill=lime] (0,0) 
	circle [radius=0.16] 
	node[white] {{\fontfamily{qag}\selectfont \tiny ID}};
	\draw[white, fill=white] (-0.0625,0.095) 
	circle [radius=0.007];
	\end{tikzpicture}
	\hspace{-2mm}
}

\foreach \x in {A, ..., Z}{%
	\expandafter\xdef\csname orcid\x\endcsname{\noexpand\href{https://orcid.org/\csname orcidauthor\x\endcsname}{\noexpand\orcidicon}}
}

\title{Using LSTM for the Prediction of Disruption in ADITYA Tokamak}

\author{
  Aman~Agarwal\thanks{Corresponding Author} \orcidA \\
  Dept. of Computer Science and Engineering\\
  Institute of Technology, \\ Nirma University\\
  Ahmedabad, IN-382481 \\
  \texttt{15bce006@nirmauni.ac.in} \\
   \And
  Aditya~Mishra\orcidB \\
  Dept. of Computer Science and Engineering\\
  Institute of Technology, \\ Nirma University\\
  Ahmedabad, IN-382481 \\
  \texttt{15bce003@nirmauni.ac.in} \\
   \And
  Priyanka~Sharma\orcidC \\
  Dept. of Computer Science and Engineering\\
  Institute of Technology, \\ Nirma University\\
  Ahmedabad, IN-382481 \\
  \texttt{priyanka.sharma@nirmauni.ac.in} \\
   \And
  Swati~Jain \\
  Dept. of Computer Science and Engineering\\
  Institute of Technology, \\ Nirma University\\
  Ahmedabad, IN-382481 \\
  \texttt{swati.jain@nirmauni.ac.in} \\
   \And
  Sutapa~Ranjan \\
  Institute for Plasma Research \\
  Gandhinagar, IN-382428 \\
  \texttt{sutapa.ranjan@gmail.com} \\
   \And
  Ranjana~Manchanda \\
  Institute for Plasma Research \\
  Gandhinagar, IN-382428 \\
  \texttt{mranjana@ipr.res.in} \\
}

\begin{document}
\maketitle

\begin{abstract}
Major disruptions in tokamak pose a serious threat to the vessel and its surrounding pieces of equipment. The ability of the systems to detect any behavior that can lead to disruption can help in alerting the system beforehand and prevent its harmful effects. Many machine learning techniques have already been in use at large tokamaks like JET and ASDEX, but are not suitable for ADITYA, which is comparatively small. Through this work, we discuss a new real-time approach to predict the time of disruption in ADITYA tokamak and validate the results on an experimental dataset. The system uses selected diagnostics from the tokamak and after some pre-processing steps, sends them to a time-sequence Long Short-Term Memory (LSTM) network. The model can make the predictions 12 ms in advance at less computation cost that is quick enough to be deployed in real-time applications.  
\end{abstract}

% keywords can be removed

%%%%%%%%%%%%%%
% INTRODUCTION
%%%%%%%%%%%%%%

\section{Introduction}
With the depletion of non-renewable sources of energy, new ways of harnessing power have come up in the form of solar energy, wind energy, hydro energy, geothermal energy, nuclear energy, and many more. One such great source of energy is plasma. Plasma is created and confined in a magnetized environment inside the vessels know as Tokamak and then used to create heat energy through the fusion of atoms and run the turbines to harness the energy. Although plasma has not yet generated enough energy to sustain human needs, it still holds a lot of promise. A major hindrance in this process is the disruption of plasma confinement. Major disruptions in plasma destroy its confinement resulting in the release of a huge amount of thermal energy which impacts the integrity of the tokamak and the surrounding pieces of equipment. Since the nature of these disruptions is highly unavoidable, a better option would be to predict its occurrence and control it.

Disruption is a violent event that occurs due to the increase in instability of the plasma and results in the termination of its magnetic confinement. A process of thermal quench and current quench starts in such cases to bring down the temperature, thereby inducing eddy and halo currents which can generate electromagnetic forces on the tokamak chamber and lead to localized heat damage or erosion. Although the machine components can withstand such events, repetitive disruption can severely reduce the vessel's lifetime \cite{Odstril2013ComparisonOA}. Therefore the detection of disruption is needed with even more increasing urgency. 

However, the prediction of this disruption is not an easy task and can be even more complex with different tokamak sizes, configurations, and physical conditions such as temperature, current, voltage, etc. Researchers have long been trying to study several diagnostic signals like the current, voltage, magnetic flux, etc, and have even applied various statistical and machine learning models for this purpose \cite{Pautasso2002OnlinePA, Lpez2012ImplementationOT, Wrblewski1997TokamakDA}.

Sengupta \emph{et al.} \cite{Sengupta2000} use Artificial Neural Network (ANN) on the past time-series values of various diagnostics of ADITYA tokamak to predict their future values thereby approximating the disruption. They were able to predict the disruption of 8 ms in advance with little distortions. Windsor \emph{et al.} \cite{Windsor2005ACN} used a cross-tokamak approach where the model was trained on one tokamak and then tested on another. JET and ASDEX were used in the experiment. Using a single layer perceptron, they were able to get a test accuracy of around 67\% when trained on one and tested on another tokamak, whereas, around 90\% when trained and tested on the same tokamak. A different approach was followed by Cannas \emph{et al.} \cite{Cannas2007DynamicNN} where they tried to predict the future values of the diagnostic signals and monitored the error in the prediction. When the error rate was low, the plasma was assumed to be stable, whereas, when the error rate increased the plasma was expected to be unstable thereby signifying that the disruption is imminent.

In \cite{Ratt2008FeatureEF}, Rattá \emph{et al.} used Support Vector Machines (SVM) to achieve a good accuracy 30 ms before the disruption, while in \cite{Ratt2012ImprovedFS} they apply Genetic Algorithm (GA) for improved feature selection and show accuracy in prediction of around 91\% 100 ms before the disruption. A similar cross-tokamak approach discussed in \cite{Windsor2005ACN} has been implemented by Kates-Harbeck \emph{et al.} \cite{KatesHarbeck2019PredictingDI} using deep learning. They utilized both Convolutional Neural Networks (CNN) and Recurrent Neural Networks (RNN) for harnessing spatio-temporal features in the prediction process. The data came from DIII-D and JET and gave the highest accuracy of 81\% on cross-machine and 95\% on single machine tests. Diogo \emph{et al.} \cite{Ferreira2018ApplicationsOD} uses a CNN to extract and filter the bolometer and plasma current signals from the tokamak and then pass it through a sequence model (LSTM) to predict the point of disruption through two methods: the probability of disruption and the time to disruption. Combining both methods and changing the prediction threshold, they show how the selection of threshold can affect the false positive rate of the prediction. They trained their model on a dataset of around 10,000 samples from JET and were able to achieve a success rate of 85\%. The advantage of using deep learning here is that the selection of features (diagnostic signals in this case) is no longer a concern since these networks themselves perform feature-engineering to find the relevant information from the input.
 
\begin{table}
 \caption{Technical specifications of ADITYA tokamak.}
  \centering
  \begin{tabular}{ll}
    \toprule
    Parameter & Value \\
    \midrule
    Major Radius & 0.75 m \\
    Minor Radius & 0.25 m \\
    Magnetic Field & 1.2 T \\
    Plasma Current & 250 kA \\
    Discharge duration & 100 ms (pulse) \\
    \bottomrule
  \end{tabular}
  \label{tab:table_technical_specification}
\end{table}

The above methods, and many others used for disruption prediction, are very hard to compare against one another since there is no standard dataset or measurement technique yet. The configuration of each tokamak is different and also the diagnostics used to measure the electrical signals can change significantly. Therefore, one method can work very well for a particular tokamak and not work at all for another. The study discussed in this paper was focused on the ADITYA tokamak at the Institute for Plasma Research in India, the technical specifications of which can be found in table \ref{tab:table_technical_specification}. As compared to ITER, ASDEX, JET, or DIII-D tokamak, ADITYA is very small and the shots last for a shorter time. Even the diagnostic signals used here are quite limited. So most of the methods discussed above are difficult to work for ADITYA.

\subsection{Proposed Approach}
 
\begin{figure*}
    \centering
    \includegraphics[width=0.7\linewidth]{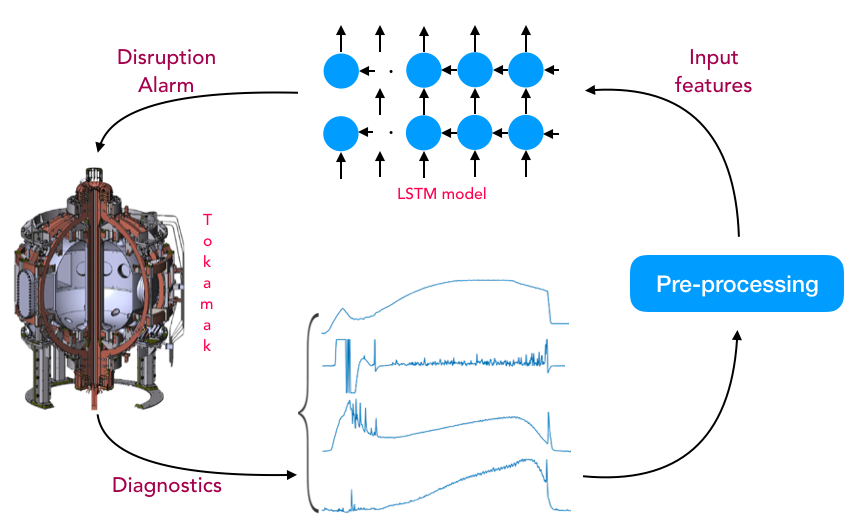}
    \caption{An flow diagram of the working of our model on a live tokamak.}
    \label{fig:overview}
\end{figure*}

We have come up with a deep learning based solution to this problem, where, using various diagnostics signals from the tokamak, we predict the disruption in plasma confinement 12 ms before it occurs. By using LSTM \cite{lstm}, a type of RNN, we were able to achieve an accuracy of 95\% on the test set with very minimal computation time. A flow diagram of the process can be seen in fig. \ref{fig:overview}.

This study differs from the others mentioned above on three grounds:
\begin{enumerate}
    \item In some of earlier studies and other state-of-the-art techniques, the prediction time is well in advance than 12 ms. While such an early prediction time interval, of more than even 100 ms, is possible for big tokamak like JET or ITER, it is not applicable on small tokamak like ADITYA, where the average duration of an entire shot is around 120 ms. 
    
    \item On ADITYA, Sengupta \emph{et al.} \cite{Sengupta2000} were able to achieve a prediction interval of 8 ms with ANN. But, as per their study, they used only one training sample (shot) and three testing samples. While they achieved good results, such a small amount of data is not enough for establishing a concrete technique for predicting disruption in a variety of shots. Further research on ADITYA tokamak in this regard is very limited.
    
    \item The prediction technique used for the real-time application should be fast enough to compute on the signals at a millisecond scale. Using stateful LSTM networks, our method could compute the signals within a millisecond interval and can be used in the real-time prediction of disruption.
\end{enumerate}

In this approach, we select a few diagnostic signals from the tokamak and run them through a series of pre-processing steps like re-sampling and normalization. These discrete, time-sequence signals are then passed through a recurrent model which predicts an output ranging from 0 to 1 at every step. The output relates to the probability of disruption at that time instance. At any step, if the output value is greater than a certain threshold, it indicates that a major disruption is going to occur. 

We have successfully validated the results of this study on a dataset of 36 test shots, finding and analyzing various pre-cursors to the disruption. Furthermore, usage of LSTM allowed input of only the current signal values instead of the series of previous values (windowing approach), as used in \cite{Sengupta2000}, and thus, our model was able to achieve a real-time prediction within 170 $\mu$s on an Intel Xeon processor which makes it easily deployable on the live tokamak. 
%Section \ref{sec_dataset} talks about the dataset 

%%%%%%%%%%%%%%%
% DATASET
%%%%%%%%%%%%%%%

\section{Dataset}
\label{sec_dataset}

The dataset used here was a set of 119 disruptive shots from the ADITYA tokamak. Out of several diagnostics, a few temporal diagnostic signals were collected to prepare the dataset. These diagnostics included:

\begin{enumerate}
    \item Plasma current (\pIP)
    \item Loop voltage (\pVLOOP)
    \item Bolometer probe (\pBOLO)
    \item One Mirnov 16$^{\circ}$ probe (\pMIRNOV)
    \item Hard X-ray (\pHXR)
    \item Soft X-ray (\pSXR)
    \item \pHA \hspace{0.2pt} monitor
    \item \pC
    \item \pDELTA
    \item \pO
\end{enumerate}

The dataset comprises of the experimental shots that were conducted on the tokamak for research purposes and all of the shots ended in a disruption. The sampling frequency of the diagnostic signals also differs based on the measuring instrument and the signal property. Most of the signals were sampled at a frequency of 0.2 ms i.e., their values were recorded after every 0.2 ms time interval. While, that of \pHXR, \pSXR, and \pMIRNOV \hspace{0.2pt} were sampled at an interval of 0.008 ms. Deep learning models usually work on the data of fixed dimensions and hence, the difference in the sampling frequency of the signals needed to be removed. This could have been achieved by either under-sampling the \pHXR, \pSXR, and \pMIRNOV \hspace{0.2pt} signals to 0.2 ms or by interpolating the rest of the signals to 0.008 ms interval. The former approach was used based on two reasons,

\begin{enumerate}
    \item Interpolating a signal is an approximation and that might have added marginal errors in the signal values, thus affecting the model accuracy.
    \item Re-sampling the signals and running the algorithm at an interval of every 0.008 ms would have increased the number of unnecessary calculations and over-burdened the processors. The interval of 0.2 ms in predicting the disruption was adequate.
\end{enumerate}

\begin{figure}
    \centering
    \includegraphics[width=0.7\linewidth]{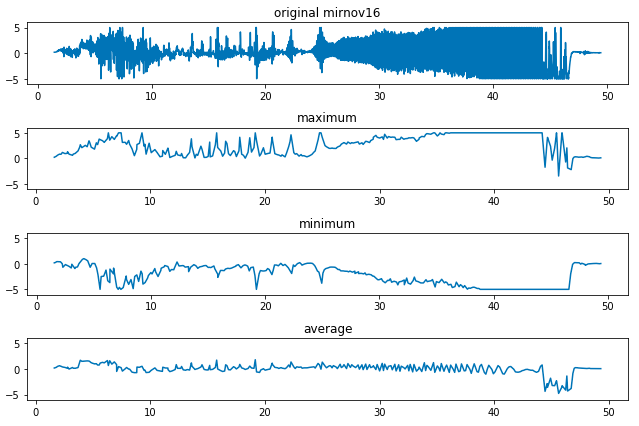}
    \caption{Down sampling of \pMIRNOV \hspace{0.2pt} probe signals by taking maximum, minimum, and average over a window of 25 time-steps.}
    \label{fig:mirnov_down_sampling}
\end{figure}

To under-sample the probe readings, a few of the techniques were taken into consideration like taking the maximum, minimum, or average of the signal readings over a sliding window of 25 time-steps. The comparison of these three techniques on the \pMIRNOV \hspace{0.2pt} probe is shown in fig. \ref{fig:mirnov_down_sampling}. Since most of the signals oscillate across the x-axis, averaging would have degraded the signal. So, we settled for the maximum as it best represented the input signal. The \pMIRNOV \hspace{0.2pt} signal in fig. \ref{fig:mirnov_down_sampling} seems to have a flat top at  5 and -5 units on the graph. That is because due to the higher amplitude of the signal at increased frequencies, it has been clipped while recording by the instrument. 

\begin{figure}
    \centering
    \includegraphics[width=0.7\linewidth]{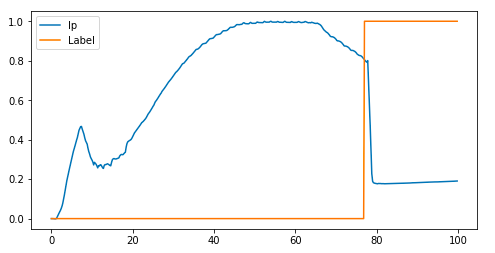}
    \caption{The label having a value of $0$ before the point of disruption and $1$ after that. The point of disruption was identified just before the \pIP \hspace{0.2pt} (scaled between 0 and 1) has a maximum fall in its signal.}
    \label{fig:label}
\end{figure}

Each shot spanned over a very short interval of time, around 100 ms, and the disruptions mostly used to occur within 30-40 ms to the start of the experiment. The sampling frequency of 0.2 ms thereby resulted in a total of 500 time-steps for the LSTM model. The next and final step was to prepare a label for the prediction of disruption. According to the theory, disruption leads to a sharp decline in the current (\pIP) signals, due to current quenching \cite{Abdullaev2015MechanismsOP}. Although other signals also show the signs, the change in \pIP \hspace{0.2pt} is significant and can be easily identified. So, the point of disruption was assumed wherever \pIP \hspace{0.2pt} had the maximum fall in its signal. The label was created just like the other signals but having a value of $0$ till the point of disruption and $1$ after the disruption. An example can be seen in fig. \ref{fig:label}. 

%%%%%%%%%
% RESULTS
%%%%%%%%%

\section{Results and Discussions}
\label{sec_res}

The network was architectured in such a way that it was neither too small to hinder the accuracy nor too large to be slow in computation. And that gave us the training accuracy of 95.39\% and test accuracy of 94.58\%. But this accuracy was not sufficient to assess the model's performance as it can't define by how much margin the model predicted the disruptions correctly. So, we calculated the difference in the time-steps of the actual and predicted disruption and divided them it into 4 categories. 

\begin{enumerate}
    \item \textbf{False alarm}, when the disruption is predicted more than 40 ms in advance than the actual disruption. At this time almost no pre-cursor is seen in the signals.
    \item \textbf{Premature alarm}, the predictions made between 40 ms and 20 ms in advance are considered premature alarms since some of the precursors start showing at this instant but still they are too early for stopping a shot.
    \item \textbf{Missed alarm}, in ADITYA at least 7-8 ms are required to implement the safety measures to prevent the harmful effects of the disruptive termination of a plasma discharge. Therefore, any prediction 8 ms after the disruption is considered as a missed alarm.
    \item The rest of the alarms which fall under the safe time limits are considered as \textbf{True alarm}.
\end{enumerate}

\begin{table}[t]
    \centering
	\caption{A summary of the model performance.}
	\begin{tabular}{l|c|c} \toprule
		\textbf{} & \textbf{Training set} & \textbf{Testing set} \\ \midrule
		\textbf{False alarm} & 2/83 (2.4\%) & 1/36 (2.8\%) \\ 
		\textbf{Premature alarm} & 18/83 (21.7\%) & 10/36 (27.7\%) \\ 
    	\textbf{Missed alarm} & 0/83 (0.0\%) & 2/36 (5.5\%) \\ 
		\textbf{True alarm} & 63/83 (75.9\%) & 23/36 (63.9\%) \\ \bottomrule
	\end{tabular}
	\label{tab:result}
\end{table}

\begin{figure}
    \centering
    \includegraphics[width=0.7\linewidth]{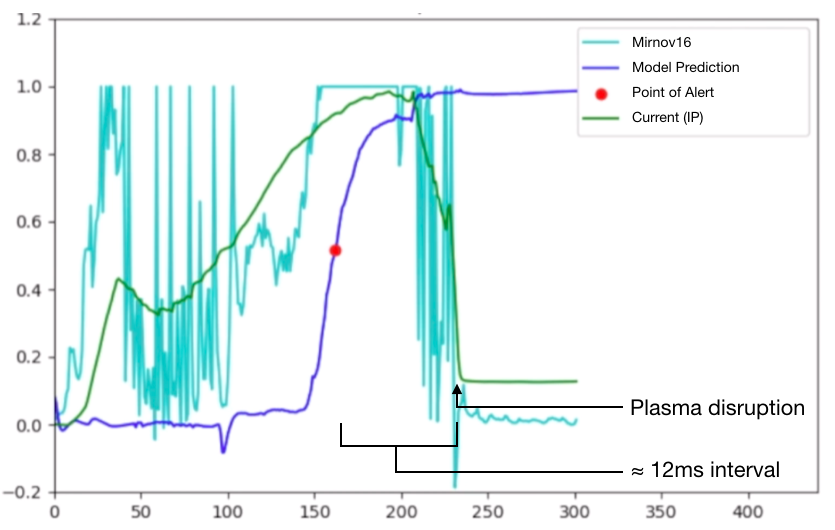}
    \caption{The plot showing the real-time prediction of disruption in an experimental shot. When the output crosses a certain threshold (say 0.5) it means the disruption is going to happen. Signals have been scaled appropriately to fit the plot.}
    \label{fig:result_real_time}
\end{figure}

Table \ref{tab:result} summarizes the results of the network on training and testing data. It is quite evident that the model performed almost similar on both the data sets. There were a few false alarms in both the training and test sets and even the missed alarm count was episodic. It can be observed that the model performed really well on the test set, correctly predicting the disruptions in 23 shots out of a total of 36 shots. Although, some work needs to be done to reduce the premature alarm count, the overall results on such a small dataset were good. See fig. \ref{fig:result_real_time} for the performance of the model in real-time\footnote{\url{https://github.com/amanbasu/plasma-disruption/blob/master/plasma_disruption.gif}}. The further details about the methods and results will be discussed in the full paper.

%%%%%%%%%%%%
% CONCLUSION
%%%%%%%%%%%%

\section{Conclusion and Future Work}
This paper discusses an approach for the prediction of disruption in the plasma confinement. Plasma is an unstable state of matter which is confined in a special vessel known as tokamak. This confinement usually ends in a violent dispersion of thermal energy, also known as a major disruption, and hampers the integrity of the vessel and its surrounding instruments. Therefore, its prediction in advance can limit such incidents and reduce the cost of its operation. Machine learning and other automatic methods have already been used which usually monitors various diagnostic features from the tokamak to predict the disruption. The main consideration while using these methods boils down to three things, (i) how quickly they process the diagnostic signals, (ii) how well in advance they predict the disruption, and (iii) by how much accuracy. While some of the studies mentioned in the paper show the prediction as early as even 1 sec with good accuracy on big tokamaks like JET and ITER, these results are not suited for small tokamak like ADITYA on which our study is based on. Even the limitation of data in ADITYA makes the experimentation more challenging. 

Through this work, it has been shown how LSTM networks can be used for the task of disruption prediction. We selected a few diagnostic signals from the ADITYA tokamak and trained the network to predict the point of disruption. The network was able to predict the results 12 ms in advance with good accuracy and the results were also validated on the test set. A prediction made this early can be used to alert the system of the imminent disruption well in advance so that preventive measures to control the thermal energy are deployed in time. Another good thing about the network is that it is real-time and can infer the results of one time-step in under 170 $\mu$s on an Intel Xeon processor running python.

One of the challenges currently with the model is the significant number of premature alarms. A possible reason might be that since there could be a lot of variations in the signal values each time the shot takes place, the model might not be able to capture every variation and thus produces an early alarm. A training data of 83 shots is still minuscule for a deep learning model which requires huge datasets to learn different varieties of the input. Perhaps in the future, more shots and diagnostics can be collected from the tokamak to prepare a more robust network.

%%%%%%%%%%%%%%%%%%
% ACKNOWLEDGEMENTS
%%%%%%%%%%%%%%%%%%

\section*{ACKNOWLEDGMENT}
This research work has been jointly collaborated with the Institute for Plasma Research (IPR), Gandhinagar. It has been financially supported by the Board of Research in Nuclear Sciences (DAE). The authors would like to thank all the scientists from IPR, who have supported directly or indirectly for understanding the domain knowledge and other details related to the project.

%%%%%%%%%%%%
% REFERENCES
%%%%%%%%%%%%

\bibliographystyle{unsrt}  
\bibliography{template}

\end{document}